\pdfoutput=1

\documentclass[11pt]{article}

\usepackage[final]{acl}

\usepackage[inkscapelatex=false]{svg}

\usepackage{booktabs}
\usepackage{amssymb}
\usepackage{pifont}
\usepackage{xcolor}
\usepackage{colortbl}
\usepackage{array}
\usepackage{xspace}
\usepackage{multirow}
\usepackage{adjustbox}
\usepackage{amsmath}
 \usepackage[utf8]{inputenc}
\usepackage{boldline}
\usepackage{microtype}
\usepackage{multicol}
\usepackage{multirow}
\usepackage{inconsolata}
\usepackage{svg}
\usepackage{graphicx}
\usepackage{listings}
\usepackage{epsfig,endnotes}
\usepackage[utf8]{inputenc}
\usepackage[T1]{fontenc}
\usepackage{upquote}
\definecolor{naturegreen}{RGB}{0, 156, 0}  
\definecolor{naturered}{RGB}{159, 0, 0}    

\usepackage{times}
\usepackage{latexsym}

\usepackage[T1]{fontenc}

\usepackage[utf8]{inputenc}

\usepackage{microtype}

\usepackage{inconsolata}

\usepackage{graphicx}

\title{Docs2Synth: A Synthetic Data Trained Retriever Framework for \\ Scanned Visually Rich Documents Understanding}
\author{
Yihao Ding\textsuperscript{1}\thanks{Equally contributed.},
Qiang Sun\textsuperscript{1}\footnotemark[1],
Puzhen Wu\textsuperscript{2},
Sirui Li\textsuperscript{3},
Siwen Luo\textsuperscript{1},
Wei Liu\textsuperscript{1}\thanks{Corresponding author.}
\\
\textsuperscript{1}University of Western Australia,
\textsuperscript{2}The University of Hong Kong,
\textsuperscript{3}Murdoch University
\\
\tt\small
\{yihao.ding,pascal.sun,siwen.luo,wei.liu\}@uwa.edu.au, \\
\tt\small
puzhenwu8@connect.hku.hk, sirui.li@murdoch.edu.au
}

\begin{document}
\maketitle
\begin{abstract}
Document understanding~(VRDU) in regulated domains is particularly challenging, since scanned documents often contain sensitive, evolving, and domain specific knowledge. This leads to two major challenges: the lack of manual annotations for model adaptation and the difficulty for pretrained models to stay up-to-date with domain-specific facts. While Multimodal Large Language Models~(MLLMs) show strong zero-shot abilities, they still suffer from hallucination and limited domain grounding. In contrast, discriminative Vision-Language Pre-trained Models~(VLPMs) provide reliable grounding but require costly annotations to cover new domains.
We introduce \textbf{Docs2Synth}, a synthetic-supervision framework that enables retrieval-guided inference for private and low-resource domains. \textbf{Docs2Synth} automatically processes raw document collections, generates and verifies diverse QA pairs via an agent-based system, and trains a lightweight visual retriever to extract domain-relevant evidence. During inference, the retriever collaborates with an MLLM through an iterative retrieval--generation loop, reducing hallucination and improving response consistency. We further deliver \textbf{Docs2Synth} as an easy-to-use Python package, enabling plug-and-play deployment across diverse real-world scenarios.
Experiments on multiple VRDU benchmarks show that Docs2Synth substantially enhances grounding and domain generalization without requiring human annotations. Our open-source implementation is available at: \url{https://docs2synth.ai4wa.com}, and the demonstration video is available at \url{https://docs2synth.ai4wa.com/video}.
\end{abstract}

\section{Introduction}

\label{sec:positional_feature}
\begin{figure}[h]
    \centering
    \tiny \includegraphics[width=\linewidth]{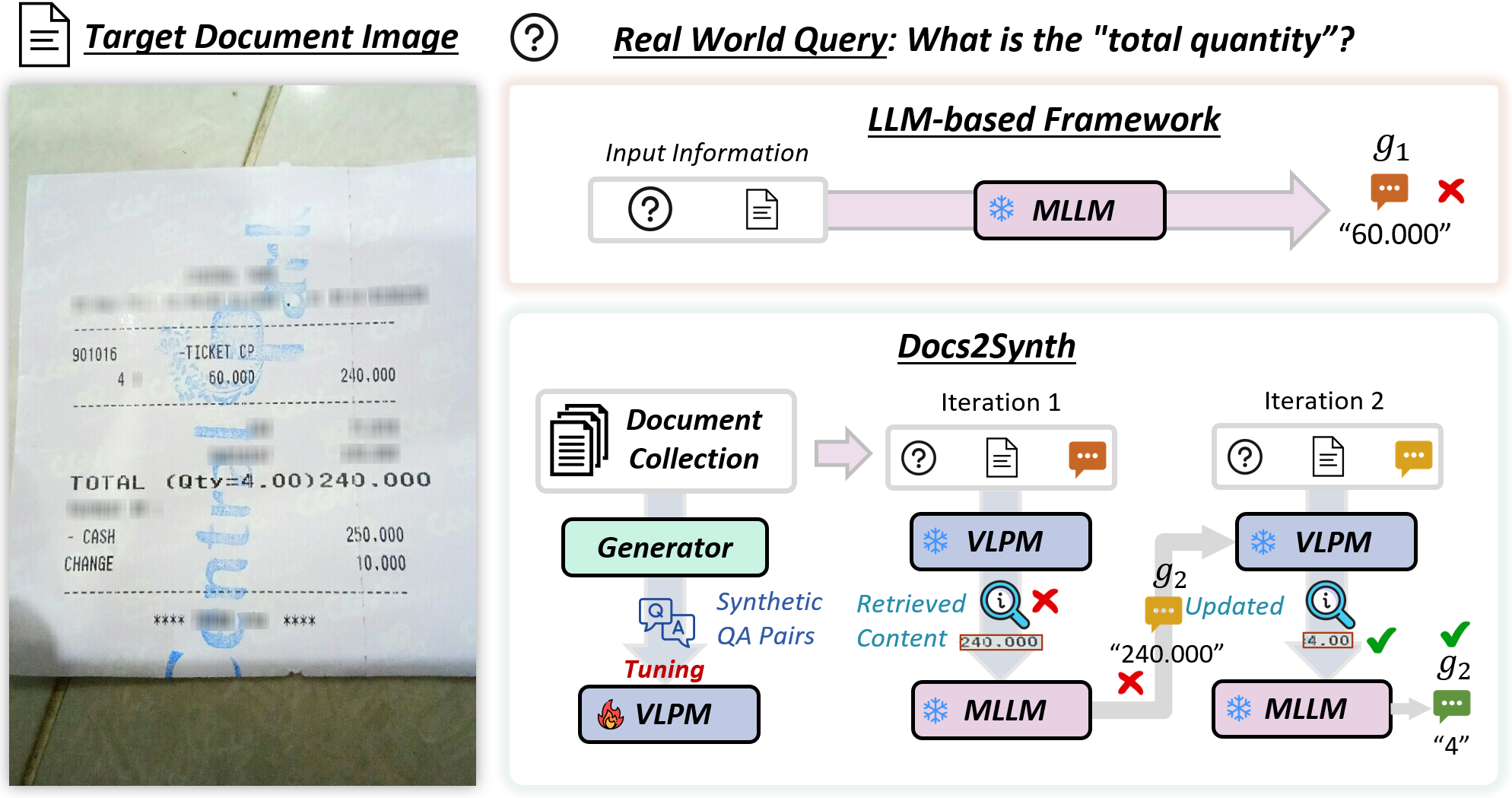}
    \caption{Typical MLLM-based VRDU and Docs2Synth.}
    \label{fig:intro}
\end{figure}  

\begin{figure*}[t]
  \centering
  \includegraphics[width=\linewidth]{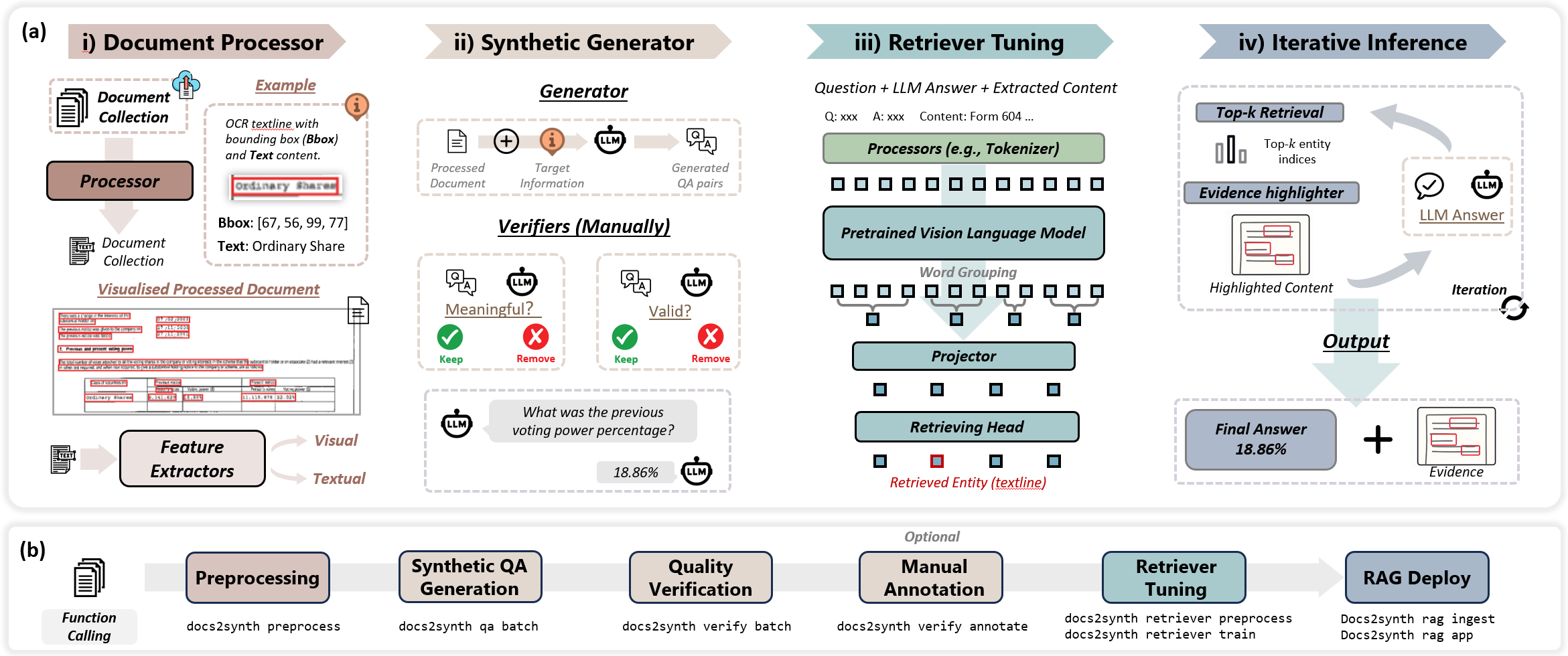}
  \caption{Design overview of the \textbf{Docs2Synth} framework.
(a) Backend architecture. (b) End-to-end data flow and function execution pipeline.}
\label{fig:backend}
\end{figure*}
The demand for automatic information extraction from visually rich documents has grown rapidly across diverse domains, including finance~\cite{formnlu}, retail~\cite{cord}, and politics~\cite{vrdu}. Multimodal Large Language Models (MLLMs)~\cite{luo2024layoutllm,gemini1.5} have been widely explored to meet this need, demonstrating promising performance and being applied in practice. However, applying these generative frameworks to knowledge-intensive tasks remains challenging due to issues such as hallucination and insufficient grounding~\cite{nourbakhsh2024towards}. These challenges are further exacerbated in scanned documents—such as handwritten forms and invoices—which introduce additional complexities due to Optical Character Recognition (OCR) inaccuracies, document skew or rotation, and handwriting variability. 

In recent years, layout-aware Vision–Language Pre-trained Models (VLPMs)~\cite{layoutlmv3,wukong}, have achieved promising performance in key information extraction~(KIE) when fine-tuned on carefully curated, task-specific datasets. 
However, the practicality of these discriminative models is limited in real-world settings, where high-quality annotations are scarce—particularly for new, domain-specific document collections.
To overcome this annotation bottleneck, recent efforts have turned to synthetic data generated by MLLMs~\cite{llava1.5,mplugdocowl2}, which can automatically produce diverse instruction–response pairs to enhance model generalization and instruction-following capabilities.
A recent study \cite{ding2025synjac} suggests that such synthetic corpora can support the domain adaptation of VLPMs under few-shot scenarios. Some VRDU-oriented MLLMs have also constructed large-scale synthetic datasets for self-supervised pretraining~\cite{feng2024docpedia} and instruction-tuning~\cite{hu2024mplug}. Nonetheless, a significant gap remains in understanding: (1) the effectiveness of fine-tuning lightweight discriminative VLPMs on large-scale, synthetic, domain-specific corpora to handle newly scanned documents without manual annotations, and (2) the potential of fine-tuned models in supporting or enhancing stronger MLLMs during inference. 

In this paper, we address the challenge of adapting VRDU systems to new, privately scanned documents without manual annotations.
We design a retrieval-guided paradigm in which a lightweight visual retriever is first adapted to the new domain through adaptive tuning, and then used to ground MLLM responses during inference, improving reliability in knowledge-intensive queries.

\noindent\textbf{Our contributions are as follows:}

i) We propose \textbf{Docs2Synth}, a novel framework that automatically digests new document collections, generates high-quality synthetic QA pairs, and trains a domain-adapted visual retriever to support retrieval-guided iterative MLLM inference, ensuring reliable and well-grounded responses in private and low-resource domains.

ii) We provide \textbf{Docs2Synth} as an easy-to-use \textbf{\textit{open source Python package}} that automates data processing, synthetic corpus creation, retriever training, and retrieval-enhanced inference, enabling scalable and practical deployment of MLLM-based VRDU across diverse real-world domains.



\section{Docs2Synth Framework}

\noindent\textit{\textbf{i) Document Processor.}} Given a new document collection $\mathbb{D}$, each document image $d \in \mathbb{D}$ is processed by an OCR tool $\mathcal{P}$ to extract a set of semantic entities $L = \mathcal{P}(d) = \{l_1, \dots, l_n\}$, where each entity $l_i$ is a pair $(c_i, b_i)$ consisting of text content $c_i$ and its bounding box $b_i$. Using XY-cut reading order methods, the full document text is aggregated as $T$ and corresponding bounding box list $B$. The entities are then passed through pretrained language and vision backbones to obtain textual features $E_T = \{E_{t_1}, \dots, E_{t_n}\}$ and visual features $E_V = \{E_{v_1}, \dots, E_{v_n}\}$.

\noindent\textit{\textbf{ii) Synthetic QA Generator.}} To reduce manual annotation efforts, we employ a MLLM $\mathcal{G}$ to create synthetic question-answer (QA) pairs based on the extracted entity set $L$. For each entity $l_i \in L$, we prompt $\mathcal{G}$ to generate a question $q_i$ such that the answer is $c_i$: $\mathcal{G}(c_i, d) \rightarrow q_i$. For example, given the context “\textit{Ordinary Shares}” in document $D$, $\mathcal{G}$ may generate the question “\textit{What is the type of shares listed in the document?”}. To ensure the quality of the QA pairs, we allow the MLLM to act as a verifier by (1) evaluating the relevance and clarity of the generated question $q_i$, and (2) confirming whether $c_i$ is a valid answer to $q_i$. Each document image yields a QA pair set $P = (Q, C)$, where $Q = \{q_1, \dots, q_m\}$ and $C = \{c_1, \dots, c_m\}$ are the generated questions and corresponding answers, with $m$ as an adjustable parameter.

\noindent\textit{\textbf{iii) Retriever Tuning.}}
To train a high-quality retriever, we fine-tune a layout-aware pretrained VRDU model $\mathcal{R}$ using synthetic QA pairs. Each training sample consists of a question $q_i$, its corresponding document image $d$, full text content $T$, bounding box list $B$, entity-level features $E_T$ and $E_V$, and an initial answer prediction $a_i$ generated by a MLLM. The retriever aims to identify the correct answer entity $l_i$ from the candidate set $L = {l_1, \dots, l_n}$ by predicting its index. Formally, $\mathcal{R}$ learns a scoring function :
\[
\mathcal{R}(q_i, D, T, B, a_i, E_T, E_V) \rightarrow \hat{y}_i \in \{1, \dots, n\},
\],
where $\hat{y}_i$ is the predicted index of the answer entity, and the model is trained by minimizing the cross-entropy loss between $\hat{y}_i$ and the ground-truth index $y_i$ corresponding to the true answer content $c_i$.

\noindent\textit{\textbf{iv) Iterative Inference.}}  
At each iteration step $t$, the model predicts a set of top-$k$ entity indices, denoted by, $\hat{Y}_i^t = \{i_1, \dots, i_k\}$.
Using these indices, we collect the corresponding text contents to form the retrieved content set, $C_i^t = \{c_j \mid j \in \hat{Y}_i^t\}$,
where each $c_j$ is the text content of entity $l_j \in L$. To visually emphasize the selected entities, their bounding boxes are highlighted in red on the document image, resulting in a modified image $D'$.

The multimodal LLM $\mathcal{F}$ then takes as input the question $q_i$, updated document image $D'$, full text $T$, and the retrieved content $C_i^t$, and generates an updated answer $a_i^t$: \begin{equation}
\mathcal{F}(q_i, D', T, C_i^t) \rightarrow a_i^t .
\end{equation}

This predicted answer is used to refine the next set of top-$k$ entity predictions via the retriever $\mathcal{R}$:
\begin{equation}
\hat{Y}_i^{t+1} = \mathcal{R}(q_i, D, T, B, a_i^t, E_T, E_V),
\quad 
\end{equation}
\begin{equation}
    |\hat{Y}_i^{t+1}| = k,\; \hat{Y}_i^{t+1} \subseteq L.
\end{equation}

The process continues iteratively. Both the number of retrieved entities $k$ and the total number of iterations can be specified by the user. The top-$k$ entities are selected based on the logits produced by the retriever model.

\begin{figure*}[t]
  \centering
  \includegraphics[width=\linewidth]{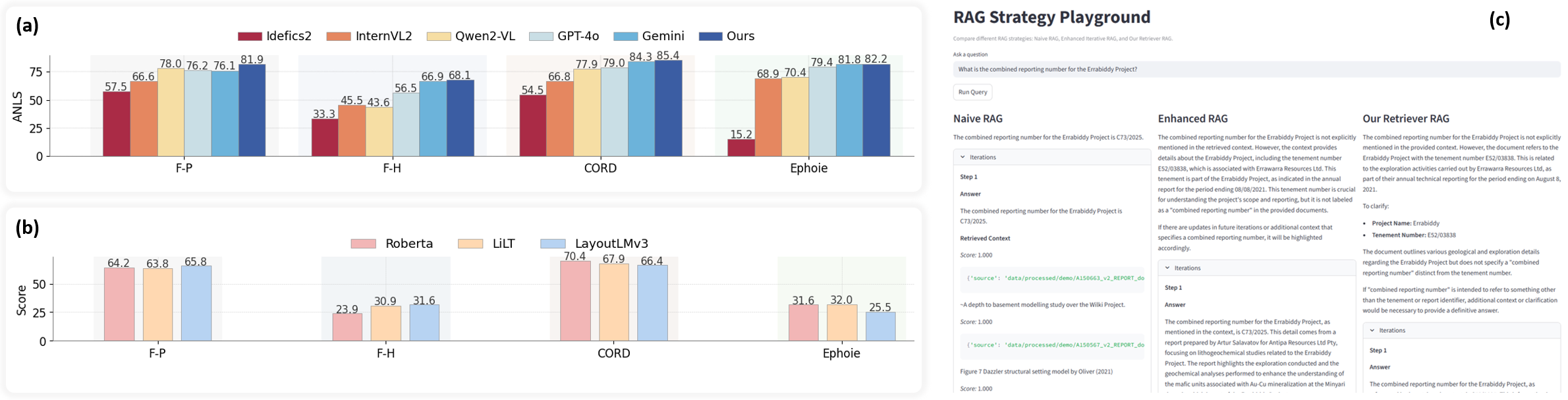}
  \caption{Quantitative results and frontend examples.}
\label{fig:results}
\end{figure*}
\section{Evaluation}
\section{Docs2Synth Implementation}

To bridge the gap between research prototypes and industrial deployment, \textbf{Docs2Synth} is delivered as a \textbf{modular}, \textbf{extendable}, and \textbf{production-ready} Python package. It enables a fully automated pipeline from document ingestion to retrieval-guided inference, supported by the following components:

    \noindent \textbf{i) Preprocessing:} Multiple OCR backends (Docling \footnote{\url{https://github.com/docling-project/docling}}, PaddleOCR\footnote{\url{https://github.com/PaddlePaddle/PaddleOCR}}, etc) are supported to extract text and layout from both scanned and digitally-generated documents. Additional OCR engines can be seamlessly integrated.

    \noindent \textbf{ii) Agent Wrapper:} A unified interface abstracts different LLM providers, allowing smooth switching between proprietary APIs~(OpenAI, Gemini, QwenVL) and local inference providers (Ollama, vLLM, HuggingFace). This design ensures scalability and extensibility to future providers while supporting deployment in privacy-sensitive environments.

    \noindent \textbf{iii) Synthetic generator:} High-quality synthetic QA pairs are generated and verified through LLM-driven agents. An optional lightweight human annotation interface further refines and supervises the generation process, improving reliability for private domains.

    \noindent \textbf{iv) Retriever Tuning:} Verified QA data is automatically transformed into training sets for lightweight VLPM retrievers. One-command training, validation, and packaging enable efficient domain adaptation.

    \noindent \textbf{v) Retrieval-Augmented Inference:} During deployment, the retriever collaborates with an MLLM under an iterative retrieval-generation strategy to ensure grounded and trustworthy responses. We also support simple RAG baselines and provide a side-by-side interface for comparing and customizing retrieval strategies, enabling users to further enhance performance.

All components are configured through a single \texttt{config.yml}, and the entire workflow—from preprocessing to production-ready inference—can be executed with a unified command (e.g., \texttt{docs2synth run}). This design ensures practical adoption in real-world systems while preserving flexibility for research and development. The system is lightweight in computation. With proprietary LLM APIs, only retriever training requires moderate resources and can run on CPU. Fully local deployment requires a GPU for inference; our experiments on an RTX 5090 (32 GB) show that the system remains feasible even with limited on-premise GPU capacity.

\subsection{Setup}
\noindent\textbf{Datasets} Several benchmark datasets support key information extraction in scanned documents. Form-NLU \cite{formnlu} targets financial forms with printed and handwritten formats, extracting 12 key fields like “Substantial Holder Name.” CORD \cite{cord} focuses on receipts, identifying fine-grained data such as “store name”. Ephoie \cite{vies} handles scanned Chinese exam headers with handwritten elements, extracting fields like “Score” and “Student Name.” 

\noindent\textbf{Implementation Details} Our system supports flexible document parsing and encoding configurations, and we release all tools and checkpoints used to produce the results in Section~\ref{sec:backend_performance}. By default, PaddleOCR is used to extract text and bounding boxes
We benchmark against strong open-source (Qwen2-VL~\citep{Qwen2-VL}, Idefics2~\citep{Idefics2}, InternVL2~\citep{internvl}) and proprietary MLLMs (GPT-4o~\citep{gpt4o}, Gemini 1.5~\citep{gemini1.5}), all evaluated under their default HuggingFace inference settings. For optional warmer tuning, we use a batch size of 16, a learning rate of $2\times10^{-5}$, and AdamW as the optimizer.
\subsection{Backend Performance}
\label{sec:backend_performance}
To demonstrate the system’s improved performance, we evaluate its effectiveness and generalizability across three benchmark datasets spanning multiple domains in Figure~\ref{fig:results}.

\noindent \textbf{Overall Performance.} As shown in Figure~\ref{fig:results} (a), the system equipped with a synthetic-data–tuned retriever consistently outperforms the best-performing MLLM across various scenarios. The improvement is particularly notable on the FormNLU-printed (F-P) set, likely due to the higher fidelity of the generated synthetic samples, which better capture the visual and structural characteristics of the printed domain. 

\noindent \textbf{Retriever Performance.} We further present the performance of the synthetic-data–tuned retriever in Figure~\ref{fig:results} (b). Overall, the tuned retriever achieves solid and reliable performance. While occasional retrieval errors remain, the consistent improvement highlights the value of incorporating synthetic data to enhance retrieval quality and downstream task accuracy. With ongoing advances in document parsing and data synthesis technologies, we expect the retriever to deliver even greater benefits for future document understanding tasks.

\noindent \textbf{Qualitative Example.}
We provide a qualitative case study in Figure~\ref{fig:results} (c) to illustrate how iterative tuning enhances both MLLM and retriever performance. Initially, the MLLM produces an incomplete student name. However, once the retriever supplies more focused and relevant content as supplementary context, the MLLM is able to generate the correct prediction. This example demonstrates the bidirectional flow of information between the MLLM and the retriever, highlighting how iterative refinement leads to mutual performance improvements.

\section{Conclusion}
We introduce a unified framework that leverages synthetic data to adaptively fine tune a domain-adapted multimodal retriever, which in turn iteratively enhances LLM inference quality on VRDU. In particular, we provide a fully packaged Python library that allows users to automatically execute the entire pipeline with minimal configuration. Experimental results demonstrate that our synthetic-data–driven adaptation strategy consistently improves downstream performance across various domains. These findings highlight the robustness, practicality, and effectiveness of the proposed approach, offering a generalizable pathway for boosting multimodal LLM performance in real-world document-centric applications.
\bibliography{custom}

\end{document}